\documentclass[11pt, a4paper, , copyright, goog]{google}

\usepackage[authoryear, sort&compress, round]{natbib}
\usepackage{xcolor}
\usepackage{amsmath,amssymb}
\usepackage{algorithm}
\usepackage{algpseudocode}
\usepackage{multirow}
\usepackage{booktabs}
\usepackage{tabularx}
\usepackage{array}

\newcolumntype{Y}{>{\centering\arraybackslash}X}
\newcolumntype{L}[1]{>{\raggedright\arraybackslash}p{#1}}

\usepackage[table]{xcolor}

\definecolor{sotabg}{gray}{0.965}
\definecolor{sotafg}{gray}{0.36}

\newlength{\sotatablewidth}

\newcommand{\sota}[1]{\textcolor{sotafg}{#1}}

\newcommand{\sotainfo}[3]{%
\sota{\textbf{Model:} #1 \hspace{0.9em}
\textbf{Metric:} #2 \hspace{0.9em}
\textbf{Score:} #3}%
}

\defcitealias{hassan2025factorized}{FVG}
\defcitealias{zheng2025vbench2}{VBench-2.0}
\defcitealias{feng2025narrlv}{NarrLV}

\defcitealias{kamath2025geneval}{GenEval2}
\defcitealias{li2025easier}{T2I-CoRe}
\defcitealias{huang2025t2i}{T2I-Comp}

\newcommand{\sotacite}[1]{%
\sota{\scriptsize \citetalias{#1}~\citeyearpar{#1}}%
}

\newcommand{\sotarefrow}[5]{%
\multicolumn{8}{@{}>{\columncolor{sotabg}}p{\sotatablewidth}@{}}{%
\begin{tabularx}{\linewidth}{@{}
>{\raggedright\arraybackslash}p{2.0cm}
>{\raggedright\arraybackslash}p{2.55cm}
>{\raggedright\arraybackslash}X
@{}}
\sotacite{#1} & \sota{#2} & \sotainfo{#3}{#4}{#5}
\end{tabularx}%
}\\[-1pt]
}
\definecolor{yscolor}{rgb}{1.0, 0.6, 0.0}

\definecolor{cldarkgreen}{rgb}{0.0, 0.6, 0.2}

\uselogo{} 

\title{PreviewDiff: Multimodal Critic-Guided Search over Diffusion Latents}

\correspondingauthor{vsub851@mit.edu}

\author[1,2, *]{Vighnesh Subramaniam}
\author[1]{Boris Katz}
\author[1,3]{Brian Cheung}
\author[2]{Chun-Liang Li}
\author[2]{Tomas Pfister}
\author[2]{Yale Song}

\affil[1]{MIT CSAIL}
\affil[2]{Google Cloud AI Research}
\affil[3]{UCSF}

\begin{abstract}
Diffusion models can produce striking images and videos, but they still struggle with the compositional details that make a generation faithful to a prompt, such as object counts, attribute binding, spatial relations, and temporally grounded actions. A common way to improve prompt satisfaction is to spend more compute at test time through Best-of-N sampling, but final-sample selection is fixed. Best-of-N can only choose among completed outputs and cannot repair a promising trajectory before it fails. We introduce \textbf{PreviewDiff}, a training-free test-time search method that turns diffusion sampling from scalar search into a multimodal critic-guided search over intermediate latents. At selected denoising checkpoints, PreviewDiff decodes a partial preview, asks a multimodal judge to score and critique it, and uses the resulting natural-language feedback to branch over semantic prompt edits and locally re-noised latent continuations. These branches are then scored and selectively rolled forward, allowing verifier compute to guide generation while the sample is still editable. Across image and video generation benchmarks, PreviewDiff consistently improves over budget-matched Best-of-N selection and strong scalar-search baselines. Ablations show that earlier interventions and increased search width provide the largest gains, while deeper search and additional semantic variants offer complementary improvements. PreviewDiff demonstrates that multimodal feedback is most useful not only as a final verifier, but as an active controller inside the denoising process.

Website: \url{https://previewdiff.github.io/}
\end{abstract}

\begin{document}

\maketitle

\section{Introduction}

Recent text-to-image and text-to-video models have achieved remarkable photorealism and broad prompt fidelity \citep{ramesh2022hierarchical, openai2024sora, flux-2-2025}. Yet despite scaling across diffusion architectures and modalities, these systems remain notoriously brittle when tasked with fine-grained compositionality \citep{chefer2023attend, ghosh2023geneval, sun2024t2vcompbench}. While standard holistic benchmarks like FID or CLIPScore often register outputs as well-aligned \citep{hessel2021clipscore, jayasumana2024rethinking}, targeted compositional evaluations consistently reveal binding errors \citep{hu2023tifa, kamath2025geneval}—such as miscounting instances, swapping attributes between nearby subjects, or misinterpreting relative spatial and temporal arrangements. Ultimately, a single feed-forward pass struggles to capture the iterative, error-correcting adjustments essential for faithful visual synthesis.

\begin{figure}
    \centering
    \includegraphics[width=\textwidth]{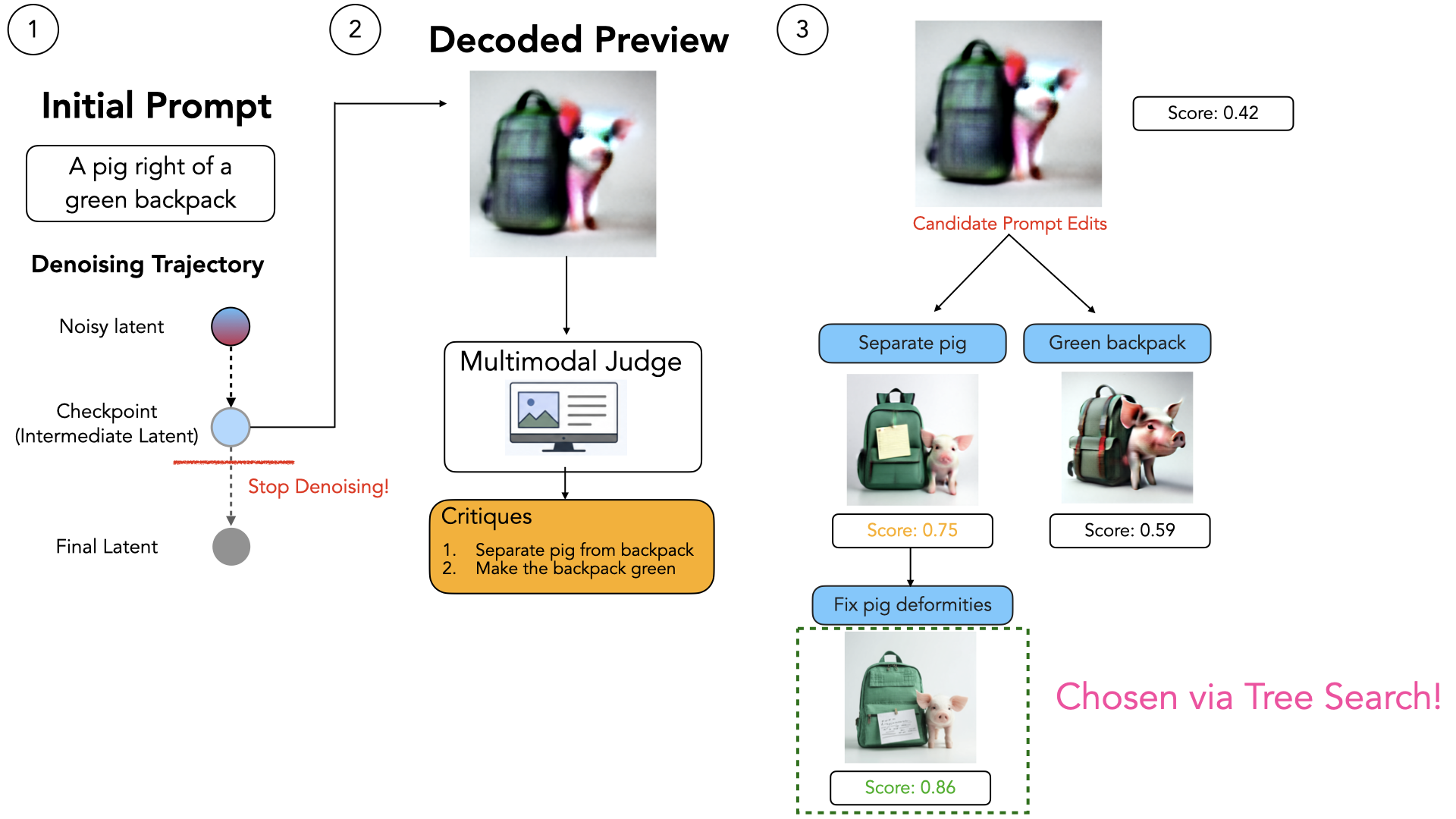}
    \caption{
    \textbf{PreviewDiff searches over critic-guided latent continuations.}
    (1) From an initial prompt, an intermediate denoising latent is decoded into a preview. 
    (2) A multimodal judge critiques the preview and proposes semantic prompt edits. 
    (3) PreviewDiff branches over these edits, scores the resulting continuations, and selects the best rollout through beam-pruned tree search.
    }
    \label{fig:concept}
\end{figure}

A natural way to improve prompt satisfaction is to spend more computation at test time. The simplest strategy is Best-of-$N$ sampling~\citep{snell2024scaling}: draw $N$ independent samples and use a reward model or multimodal judge to select the best final output. Best-of-$N$ is easy to implement and is therefore an important baseline for test-time scaling. However, it is fundamentally open-loop. It only evaluates completed samples, cannot intervene when a partially denoised sample is promising but flawed, and discards the computation spent on generations that fail late in the process. In contrast, the denoising trajectory itself contains intermediate visual evidence about whether the sample is moving toward or away from the prompt, as illustrated by the decoded checkpoint preview in Figure~\ref{fig:concept}. This suggests a different use of test-time compute. Rather than spending all verifier calls on final outputs, we can use some of them earlier to diagnose and steer partial generations.

Recent work suggests that test-time feedback can improve visual generation. Self-correcting LLM-controlled Diffusion \citep{wu2024self} closes the loop by detecting errors in generated images and applying LLM-directed latent correction operations, but this work is organized as an iterative correction loop around completed images. Other methods formulate inference-time scaling as search over sampling noises, noise trajectories, particles, diffusion trees, or evolutionary mutations \citep{ma2025inference,ramesh2025test,jain2025diffusion,singhal2025general,kim2025test,he2025scaling}. Closest to our video setting, Video-T1 \citep{liu2025videot1} frames video generation as verifier-guided test-time search and introduces tree methods to adaptively expand and prune autoregressive frame continuations. These methods show that visual generation benefits from inference-time search. Yet they generally use the verifier as a reward oracle for selecting candidates. PreviewDiff is complementary. We use a multimodal judge not only to score partial generations, but also to produce interpretable semantic actions that steer diffusion itself.

We propose \textbf{PreviewDiff}, a training-free test-time search method for improving compositional prompt following in pretrained image and video diffusion models. PreviewDiff treats intermediate denoised latent estimates as search states. At selected denoising checkpoints, a multimodal model inspects a decoded preview, estimates semantic prompt satisfaction, and produces targeted natural-language feedback about visual components of the generation. These can include missing objects, incorrect attributes, wrong counts, violated spatial relations, or temporal inconsistencies. This feedback defines semantically meaningful search actions through alternative correction prompts, local re-noising restarts, and latent continuations. The resulting branches are scored, pruned, and selectively rolled forward, turning a single open-loop denoising trajectory into a closed-loop search process guided by multimodal critique.

This design differs from both final-sample selection and scalar reward-guided diffusion. Unlike Best-of-$N$, PreviewDiff can intervene before a sample is complete and allocate additional computation to promising partial generations. Furthermore, unlike methods that steer diffusion using scalar rewards, low-level noise trajectory search, particle resampling, or terminal reward propagation \citep{ramesh2025test,singhal2025general,kim2025test,jain2025diffusion,he2025scaling}, PreviewDiff uses the multimodal judge in two roles as a value estimator for prompt satisfaction and as a policy-like source of corrective semantic actions. Unlike tree-based methods like Video-T1~\citep{liu2025videot1}, which search over video candidates and frame continuations, PreviewDiff searches over corrected diffusion latent continuations. The tree branches over interpretable object, attribute, relation, and action-level fixes rather than only over undifferentiated noise, particle, or frame candidates.

Our central empirical question is whether multimodal feedback is more useful when applied during generation than when reserved for final selection. We compare PreviewDiff against budget-matched Best-of-$N$ baselines in which the same number of multimodal-judge calls is used to select among independently generated final outputs. We also compare against scalar-search baselines such as EvoSearch \citep{he2025scaling} and particle sampling inspired by Feynman--Kac and sequential Monte Carlo steering \citep{singhal2025general}, and include Video-T1~\citep{liu2025videot1} as a video test-time scaling baseline. We evaluate scaling along several axes: semantic branching, search width, total verifier budget, checkpoint timing, tree depth, and base model scale. We show a detailed overview in Figure~\ref{fig:concept}.

In summary, this paper makes the following contributions:
\begin{enumerate}
    \item We introduce PreviewDiff, a training-free framework that casts image and video diffusion generation as multimodal, semantically guided tree search over intermediate denoising states.
    \item We use a multimodal model not only to score partial generations, but also to propose natural-language corrective actions that define search branches.
    \item We compare guided intermediate search against budget-matched Best-of-$N$ selection, scalar trajectory-search baselines, and video-specific test-time scaling methods.
    \item We demonstrate robustness across image and video generation tasks, multimodal verifiers, and downstream diffusion backbones.
\end{enumerate}

\section{Related Work}

\textbf{Inference-time search and steering for diffusion models.}
A growing line of work studies how to spend additional computation during diffusion sampling rather than only increasing the number of denoising steps. Inference-time scaling frameworks search over initial noises or sampling trajectories using verifiers and search algorithms \citep{ma2025inference,ramesh2025test}. EvoSearch formulates image and video generation as evolutionary search over denoising trajectories \citep{he2025scaling}. Tree- and particle-based methods push this idea further. Monte Carlo Tree Diffusion expands and prunes partially denoised trajectories for planning \citep{yoon2025monte}, Diffusion Tree Sampling reuses computation by propagating terminal rewards through a diffusion tree \citep{jain2025diffusion}, and Feynman--Kac steering and Diffusion Alignment as Sampling use particle resampling or sequential Monte Carlo to sample from reward-aligned distributions \citep{singhal2025general,kim2025test}. Diffusion Forcing is also related in its use of flexible intermediate uncertainty and guided sequence rollouts \citep{chen2024diffusion}. PreviewDiff shares the adaptive inference-time view, but uses the multimodal model as more than a scalar reward. More specifically, our method scores partial previews and proposes semantic correction directions that become branches over latent continuations. Video-T1 \citep{liu2025videot1} is the closest video-generation prior to our work. It also frames test-time scaling for video generation as search over trajectories from Gaussian noise to videos, using test-time verifiers and heuristic search algorithms to guide candidate selection. The Tree-of-Frames method adaptively expands and prunes autoregressive frame continuations, with verifier feedback used to balance quality and compute. PreviewDiff shares the goal of adaptive inference-time allocation, but differs in the search state and action space. Our states are intermediate denoised latent estimates, and our branches are critic-proposed semantic prompt corrections combined with local latent re-noising and continued denoising. Thus the multimodal model is not only a verifier for pruning candidates, but also a policy-like source of semantic actions.

\textbf{Visual feedback, prompt refinement, and semantic correction.}
Several systems use generated images as feedback for improving prompt adherence. Self-correcting LLM-controlled Diffusion closes a loop around completed images by detecting prompt violations and applying LLM-controlled latent correction operations \citep{wu2024self}. VisualPrompter identifies absent concepts in generated images and rewrites prompts at an atomic semantic level \citep{wu2026visualprompter}. Test-time Prompt Refinement uses an MLLM to detect missing objects or incorrect attributes and generate refined prompts for subsequent generations \citep{khan2025test}. Iterative Prompt Relabeling uses repeated image sampling and prompt relabeling to improve text-image alignment \citep{chen2024learning}. Attention- and program-based approaches such as Attend-and-Excite and VPGen/VPEval also address compositional failures including object neglect, counting, and spatial layout \citep{chefer2023attend,cho2023visual}. PreviewDiff is closest to the feedback-driven methods, but moves the feedback inside denoising. Instead of restarting a full generation after a final-image critique, it decodes intermediate clean-latent estimates, branches over critic-proposed prompt edits plus local latent restarts, and prunes before the sample is complete.

\textbf{Verifier-based selection and compositional evaluation.}
Best-of-$N$ selection is a strong baseline for test-time scaling because it can improve outputs using only a final-stage verifier or reward model. Human-preference rewards such as ImageReward and PickScore provide ranking signals for text-to-image candidates \citep{xu2023imagereward,kirstain2023pick}. VQAScore shows that VQA-style alignment scores can improve generation by selecting among multiple candidates \citep{lin2024evaluating}, and UniGen uses chain-of-thought verification for Best-of-$N$ test-time generation in a unified multimodal model \citep{tian2025unigen}. At the same time, compositional evaluation work shows why scalar final scores are often insufficient. GenEval and GenEval 2 focus on object-centric counts, colors, positions, and benchmark drift \citep{ghosh2023geneval,kamath2025geneval}. TIFA, DSG, and VQAScore decompose prompt faithfulness into question-answering or graph-structured checks \citep{hu2023tifa,cho2024dsg,lin2024evaluating}. T2I-CompBench++ and T2I-CoReBench stress attribute binding, spatial relations, numeracy, complex composition, and reasoning \citep{huang2025t2icompbenchplusplus,li2026easier}. PreviewDiff uses these verifiers in a different role. Rather than only ranking finished outputs, the same multimodal signal diagnoses partial failures and turns them into interpretable search actions.

\section{Methods}

PreviewDiff is a training-free test-time search procedure for improving compositional prompt following in pretrained image and video diffusion models. The method turns a single open-loop denoising trajectory into a tree of critic-guided latent continuations. At intermediate checkpoints, PreviewDiff decodes a partial clean-latent estimate, asks a multimodal critic to score and critique the preview, and uses the critique to branch over semantic prompt edits and local latent restarts.

\subsection{Preliminaries}
Let $p$ denote the user prompt and let $G_\theta$ be a conditional diffusion or flow model operating on latent variables. We use the standard diffusion convention in which $t=0$ denotes clean data and larger $t$ denotes higher noise. Reverse denoising therefore moves from a high-noise time $t$ toward a lower-noise time $s<t$. We write a sampler transition as
\begin{equation}
    z_s = F_\theta(z_t,c,t\!\rightarrow\!s;\epsilon), \qquad s<t,
\end{equation}
where $F_\theta$ is the transition map induced by the pretrained model, numerical sampler, scheduler, conditioning $c$, and optional sampler noise $\epsilon$. We denote the intervention checkpoints by
\[
    \mathcal{C}=(\tau_0,\tau_1,\ldots,\tau_D),
    \qquad
    T\ge \tau_0>\tau_1>\cdots>\tau_D\ge 0,
\]
where each $\tau_d$ is a diffusion time and $d$ is only the index of the checkpoint in the search schedule. In classifier-free guidance \citep{ho2022classifier}, we represent the conditioning as a positive prompt $c^+$ and a negative prompt $c^-$. Let $m_\theta$ denote the model prediction, which is a noise prediction for DDPM/DDIM-style samplers and a velocity prediction for flow-matching samplers. The guided prediction is
\begin{equation}
    m_\theta^{\mathrm{cfg}}(z_t,t,c^+,c^-)
    =
    m_\theta(z_t,t,c^-)
    +
    w\bigl(
    m_\theta(z_t,t,c^+)
    -
    m_\theta(z_t,t,c^-)
    \bigr),
\end{equation}
where $w$ is the guidance scale. When a model does not use a negative prompt, we set $c^-=\emptyset$.

At a checkpoint time $t$, the sampler exposes a clean-latent estimate $\hat{x}_{0,t}$. We assume this latent estimate to be the standard one-step clean estimate implied by the current latent and model prediction. For DDPM/DDIM-style samplers \citep{ho2020denoising,song2020denoising} with cumulative noise level $\bar\alpha_t$, this estimate is
\begin{equation}
    \hat{x}_{0,t}
    =
    \frac{
    z_t-\sqrt{1-\bar\alpha_t}\,
    \epsilon_\theta^{\mathrm{cfg}}(z_t,t,c^+,c^-)
    }{
    \sqrt{\bar\alpha_t}
    }.
\end{equation}
For flow-matching samplers with noise level $\sigma_t$ and velocity prediction $v_\theta^{\mathrm{cfg}}$, we use
\begin{equation}
    \hat{x}_{0,t}
    =
    z_t-\sigma_t v_\theta^{\mathrm{cfg}}(z_t,t,c^+,c^-).
\end{equation}
The preview shown to the critic is then
\begin{equation}
    y_t=D(\hat{x}_{0,t}),
\end{equation}
where $D$ is the image or video decoder.

\subsection{The PreviewDiff Framework}
PreviewDiff builds a search tree over intermediate denoising states. A node $n$ stores
\begin{equation}
    s_n=
    \bigl(
    \hat{x}_0^n,\,
    t_n,\,
    c_n^+,\,
    c_n^-,\,
    h_n
    \bigr),
\end{equation}
where $t_n\in\mathcal{C}$ is the diffusion time of the checkpoint associated with node $n$, $\hat{x}_0^n$ is the stored clean-latent estimate at that checkpoint, $c_n^+$ and $c_n^-$ are the current positive and negative prompt conditionings, and $h_n$ is the history of previous critic feedback. In the notation of Section~3.1, if node $n$ lies at checkpoint time $t_n$, then $\hat{x}_0^n=\hat{x}_{0,t_n}$. The decoded preview is $y_n=D(\hat{x}_0^n)$.

A multimodal critic $\mathcal{J}$ is used in two roles. First, it acts as a value model by scoring the decoded preview against the original prompt $p$. Second, it acts as a semantic proposal model by producing short natural-language corrections. These corrections target errors such as missing objects, wrong counts, incorrect attributes, broken spatial relations, and temporal or action mismatches. In our setting, rather than rewriting the full prompt, the critic proposes compact fix notes that are appended to the conditioning. Compact fix notes can preserve the original user prompt and reduce prompt drift. Full prompt rewriting can unintentionally remove constraints that were already satisfied in the partial generation. By appending only the critic's local correction, PreviewDiff keeps the original scene specification fixed while giving the sampler a targeted direction for the next continuation \citep{khan2025test}.

A branch action is
\begin{equation}
    a=(\Delta c^+,\Delta c^-,r,\xi),
\end{equation}
where $\Delta c^+$ is a positive fix note, $\Delta c^-$ is an optional negative avoid note, $r$ is a restart depth, and $\xi$ is the re-noising seed. The child conditioning is
\begin{equation}
    c_{n'}^+=c_0^+\oplus \mathrm{recent}(\Delta c^+), \qquad
    c_{n'}^-=c_0^-\oplus \mathrm{recent}(\Delta c^-),
\end{equation}
where $\oplus$ denotes prompt concatenation and $\mathrm{recent}(\cdot)$ keeps only the most recent fix notes. In our approach, the positive prompt receives the critic's requested fix, while the negative prompt receives avoid-style fragments derived from the critique. If a model does not support negative prompting, the second update is omitted.

From a parent node $n$, PreviewDiff re-noises the stored clean estimate $\hat{x}_0^n$ to a nearby higher-noise time. Let
\begin{equation}
    \rho(n,r)=\operatorname{Restart}(t_n,r),
    \qquad
    t_n < \rho(n,r)\le T,
\end{equation}
where $r$ is the restart depth and $\operatorname{Restart}(t_n,r)$ returns the scheduler time obtained by moving $r$ sampler steps backward along the reverse denoising trajectory. Thus $\rho(n,r)$ is slightly noisier than the current checkpoint time $t_n$, but it is not pure noise.

For DDIM-style samplers, this local restart is
\begin{equation}
    \tilde z_{\rho}
    =
    \sqrt{\bar\alpha_{\rho}}\,\hat{x}_0^n
    +
    \sqrt{1-\bar\alpha_{\rho}}\,\xi,
\end{equation}
and for flow-matching samplers it is
\begin{equation}
    \tilde z_{\rho}
    =
    (1-\sigma_{\rho})\hat{x}_0^n+\sigma_{\rho}\xi.
\end{equation}
The branch is then continued to the next lower-noise checkpoint $t'<t_n$ under the corrected conditioning:
\begin{equation}
    z_{t'}=
    F_\theta(
    \tilde z_{\rho},
    (c_{n'}^+,c_{n'}^-),
    \rho\!\rightarrow\!t'
    ),
    \qquad
    t'<t_n<\rho.
\end{equation}

The re-noising step is part of the search action, but we still restrict its range. In principle, a very large tree search could treat checkpoint timing, restart depth, and branch noise as unconstrained actions. In practice, each expansion requires a decode and a multimodal-critic call, so the search must operate over a finite and stable action set. PreviewDiff therefore uses local restarts: it moves the stored clean estimate $\hat{x}_0^n$ only a small number of sampler steps back to a nearby higher-noise time, then continues under the corrected conditioning. This keeps the branch near the current partial scene while giving the sampler enough freedom to realize the semantic correction. Very early checkpoints and aggressive restarts can make $\hat{x}_0^n$ unreliable, so we bound the checkpoint schedule and restart depths for both efficiency and stability. The checkpoint and search-axis ablations study how performance changes as these finite search choices vary.

Very early clean-latent estimates can be noisy or inaccurate, and aggressive re-noising can destabilize the sample. PreviewDiff mitigates this by decoding only at checkpoints where semantic structure is visible, using small restart depths, and pruning unstable children with the multimodal critic before allocating more compute. We include an ablation on checkpoint timing that probes the related tradeoff between early editability and preview reliability, while a restart-depth ablation would isolate the effect of the local restart itself.

\subsection{Beam-pruned tree search over latent continuations}
PreviewDiff adapts Monte Carlo tree search to diffusion sampling, with a beam-pruned tree policy suited to expensive visual rollouts. The search state is the node state $s_n$ defined above. The action space is the set of critic-proposed semantic corrections crossed with restart depths and latent branch seeds,
\begin{equation}
    \mathcal{A}(s_n)
    =
    \{
    (\Delta c_m^+,\Delta c_m^-,r,\xi_b)
    :
    m\le M,\,
    r\in R,\,
    b\le L
    \},
\end{equation}
with an optional no-op branch that continues without adding a new correction. Here $M$ is the number of semantic variants proposed by the critic, $R$ is the finite set of allowed local restart depths, and $L$ is the number of latent re-noising seeds per correction. The bounds on $R$ and $L$ are practical search-budget choices: expanding them increases the action space, but also increases decode and critic cost.

The reward for a node is the critic score of its decoded preview:
\begin{equation}
    R(n)=\mathcal{J}(D(\hat{x}_0^n),p).
\end{equation}

In our experiments, $\mathcal{J}$ returns a score on a simpler rubric than what is used for final evaluation. The score measures object and count correctness, attribute binding, spatial or temporal relation correctness, and overall semantic faithfulness without asking for specific details.

The search proceeds through four tree-search operations.

\textbf{Selection.}
At each checkpoint depth, PreviewDiff selects the current frontier $\mathcal{F}$ of promising nodes. We use a beam tree policy rather than a visit-count UCT policy because each child requires decoding and a multimodal-judge call. The frontier is
\begin{equation}
    \mathcal{F}=\operatorname{TopB}\{V(n):n\in\mathcal{U}\},
\end{equation}
where $\mathcal{U}$ is the union of all child nodes generated from the current frontier and $B$ is the beam width.

\textbf{Expansion.}
For each selected node, the critic proposes $M$ semantic corrections. PreviewDiff combines these corrections with restart depths and latent seeds, then creates child latent continuations using the re-noising rule above.

\textbf{Rollout.}
Each child is denoised only to the next checkpoint, decoded into a preview, and scored by the critic. At the final depth, surviving nodes are rolled out to completed images or videos and scored again.

\textbf{Backup.}
For path selection, values are propagated through the tree by a max backup,
\begin{equation}
    V(n)
    =
    \max
    \Bigl(
    R(n),
    \max_{m\in\operatorname{children}(n)} V(m)
    \Bigr),
\end{equation}
with $V(n)=R(n)$ for unexpanded leaves. Since our main approach expands each action once and then prunes, this backup reduces to selecting the best child continuation at each checkpoint. The selected final output is the highest-scoring completed rollout among the surviving leaves. 

States are intermediate clean-latent previews, actions are semantic prompt corrections plus local latent restarts, rewards are multimodal prompt-satisfaction scores, expansion creates corrected denoising continuations, and backup tracks the best descendant value. This formulation preserves the main purpose of MCTS, which is adaptive allocation of test-time compute to promising branches, while remaining practical for diffusion models where every expansion requires a costly visual decode and critic evaluation. The final algorithm is presented at Algorithm~\ref{alg:PreviewDiff}.

\begin{algorithm}[t]
\caption{PreviewDiff}
\label{alg:PreviewDiff}
\begin{algorithmic}[1]
\Require prompt $p$, generator $G_\theta$, decoder $D$, critic $\mathcal{J}$, checkpoints $\mathcal{C}$, beam width $B$, root seeds $S$, semantic variants $M$, restart depths $R$, latent branch factor $L$
\State Initialize frontier $\mathcal{F}$ by denoising each root seed $s\in S$ to the first checkpoint, decoding $\hat{x}_0$, and scoring the preview with $\mathcal{J}$.
\State $\mathcal{F}\leftarrow\operatorname{TopB}(\mathcal{F})$.
\For{checkpoint transition $\tau_d\rightarrow \tau_{d+1}$}
    \State $\mathcal{U}\leftarrow\emptyset$
    \For{node $n\in\mathcal{F}$}
        \State Decode $y_n=D(\hat{x}_0^n)$.
        \State Ask $\mathcal{J}$ to score $y_n$ and propose $M$ semantic corrections.
        \For{correction $(\Delta c_m^+,\Delta c_m^-)$, restart depth $r\in R$, and latent seed $\xi_b$ for $b=1,\ldots,L$}
            \State Form child conditioning $(c_{n'}^+,c_{n'}^-)$ by appending the correction notes.
            \State Set restart time $\rho=\operatorname{Restart}(t_n,r)$.
            \State Re-noise the stored clean latent: $\tilde z_{\rho}=Q_{\rho}(\hat{x}_0^n;\xi_b)$.
            \State Denoise $\tilde z_{\rho}$ to checkpoint $\tau_{d+1}$ under $(c_{n'}^+,c_{n'}^-)$.
            \State Decode and score the child preview with $\mathcal{J}$.
            \State Add the child node to $\mathcal{U}$.
        \EndFor
    \EndFor
    \State Backup child values to their ancestors.
    \State $\mathcal{F}\leftarrow\operatorname{TopB}(\mathcal{U})$.
\EndFor
\State Roll surviving nodes to final outputs, score them with $\mathcal{J}$, and return the best sample.
\end{algorithmic}
\end{algorithm}

\section{Experiment Setup}
\paragraph{Models.} SDXL is the main image model used for scaling experiments because its sampling path exposes intermediate clean-latent estimates and supports controlled local restarts. 
SD-3.5 and FLUX.2 are used for fixed-budget transfer experiments with stronger image backbones. 
For video generation, we evaluate LTX-Video-2.3 (9B) and Wan 2.2 \citep{hacohen2024ltxvideo,wan2025open}.
LTX-Video is the main video model used for scaling experiments, while Wan 2.2 is included as a stronger fixed-budget video backbone. 
All methods use the same base model weights within each comparison group. 
No model finetuning is performed.

\paragraph{Datasets.} For image generation, we evaluate on GenEval2, T2I-CompBench++, and T2I-CoReBench \citep{kamath2025geneval,huang2025t2i,li2025easier}. 
GenEval2 is our main image scaling benchmark because it emphasizes object-centric compositional prompt following, including counts, colors, attributes, relations, and multi-object constraints. 
T2I-CompBench++ and T2I-CoReBench are used for broader fixed-budget evaluation across compositional and reasoning-oriented prompts. 
For video generation, we evaluate on NarrLV, T2V-CompBench, and VBench 2.0 \citep{feng2025narrlv,sun2024t2vcompbench,zheng2025vbench2}. 
NarrLV is our main video scaling benchmark because it stresses narrative and temporally structured prompt satisfaction. 
T2V-CompBench and VBench 2.0 provide complementary coverage of text-video alignment, motion and action binding, temporal consistency, intrinsic faithfulness, and visual quality.

\subsection{Multimodal Evaluation}
Our primary automatic evaluator is a Gemini-based multimodal rubric using Gemini 3.1 Flash-Lite (\texttt{gemini-3.1-flash-lite}) \citep{googledeepmind2026gemini31flashlite}. 
For every generated image or video, the evaluator receives the target prompt and generated media, then assigns four binary criteria covering object and count correctness, attribute binding, relation or action correctness, and overall semantic faithfulness. 
The total score ranges from 0 to 4. 
We report the mean score under this rubric, and report exact-score-4 pass rates when appropriate. 
Unless otherwise stated, all search methods and baselines use the same Gemini model and the same rubric. We include all prompts used in the Appendix Section~\ref{ap:prompts}.

We also report auxiliary non-Gemini evaluation when available. 
For T2I-CompBench++, we use BLIP VQA over disentangled object and attribute questions \citep{li2022blip}. 
For additional image and video checks, we evaluate selected runs with Qwen3-VL-8B-Instruct \citep{bai2025qwen3vl}. 
These auxiliary evaluators are not used to guide the main Gemini search unless explicitly stated. 
They test whether gains from PreviewDiff transfer beyond the critic used during search.

To further validate our findings, we also include a human evaluation where human raters are asked to rate the best generated image or video among three approaches, PreviewDiff (Ours), Best-of-N, and EvoSearch. Our human evaluation includes nine human raters. We report three-way preference of each method as well as the pairwise win rate between PreviewDiff and EvoSearch and PreviewDiff and Best-of-N. For three way preferences, we calculate error bars as bootstrapped confidence intervals on each sample across our human raters, averaged over all ratings. Raters are asked to compare on both content (how close is the generated image or video to the prompt) and style (general aesthetic of the generated image or video) of each generated image and video to separate whether our improvements are driven by content improvements for better prompt adherence or stylistic improvements. We include rater recruitment criteria, instructions given to raters, and our evaluation UI in Appendix Section~\ref{ap:human_eval}.

For human evaluation, we use SDXL generations on GenEval2 and LTX-Video generations on NarrLV. We use budget-matched settings across PreviewDiff, EvoSearch, and Best-of-$N$ to keep the comparison fair. Raters are shown outputs in random order with method names hidden to reduce ordering and method bias. We evaluate 360 image prompts and 200 video prompts, with 9 independent ratings per prompt and criterion.

\subsection{Baselines}
\textbf{Base model.}
The base model baseline samples a single image or video from the pretrained generator using the original prompt. It uses no verifier calls during generation. This measures the unscaled prompt-following performance of each backbone. 

\noindent\textbf{Best-of-$N$.}
Best-of-$N$ samples $N$ independent completed outputs and uses the multimodal critic to select the best final sample \citep{snell2024scaling}. This is the simplest test-time scaling baseline. It spends verifier compute on final selection, but cannot intervene in a promising partial generation before the sample is complete. We include two matched settings. In the verifier-call-matched setting, $N$ equals the number of critic calls used by PreviewDiff. In the forward-pass-matched setting, $N$ is chosen so that complete Best-of-$N$ samples use approximately the same number of denoising step-equivalents as PreviewDiff. 

\noindent\textbf{EvoSearch} \citep{he2025scaling}.
EvoSearch maintains a population of latent particles along the denoising trajectory. At scheduled checkpoints, particles are rolled out to completed media and scored by the same Gemini rubric. The method preserves elites, selects parents, and produces offspring through denoising-aware latent mutations. EvoSearch therefore searches over latent trajectories, but it uses the critic only as a scalar reward. It does not ask the critic to propose semantic correction actions.

\noindent\textbf{Particle sampling} \citep{he2025scaling}.
We implement a particle-sampling baseline inspired by Feynman--Kac and sequential Monte Carlo steering. Like EvoSearch, it maintains a population of latent particles and evaluates them at intermediate checkpoints. Unlike EvoSearch, it does not mutate selected parents. Instead, particles are resampled according to a softmax over critic scores. This isolates the effect of scalar reward-based filtering along the denoising trajectory without semantic correction.

\noindent\textbf{Video-T1} \citep{liu2025videot1}.
We include Video-T1 as a video-only test-time scaling baseline. Video-T1 frames video generation as a search problem and uses test-time verifiers to guide candidate video trajectories. We compare to its reported search setting when available, including Tree-of-Frames-style adaptive expansion and pruning. This baseline is especially relevant because it also studies inference-time search for video generation, and it branches over video candidates rather than critic-proposed semantic prompt edits applied to intermediate diffusion latents.

\subsection{Scaling and Ablations}
We organize the main scaling experiments around total test-time compute, search width, semantic branching, checkpoint timing, and tree depth. Unless otherwise stated, image scaling is performed with SDXL on GenEval2, and video scaling is performed with LTX-Video on NarrLV.

We report compute in terms of denoiser steps and verifier calls rather than wall-clock latency. This distinction matters because Best-of-$N$ samples can be generated and scored in parallel, while PreviewDiff has sequential dependencies across search depths. For PreviewDiff, increasing the budget scales combinations of root seeds, beam width, semantic correction variants, latent branch factor, and tree depth. For Best-of-$N$, increasing the budget increases the number of complete samples selected by the verifier.

For ablations, we vary one axis at a time. Checkpoint timing changes when intermediate previews are decoded and judged. Tree depth changes the number of sequential judge-and-branch rounds. Semantic variants changes the number of critic-proposed prompt fixes considered at each expansion. Search width changes the number of root trajectories and partial latent states retained by the beam. We also evaluate checkpoint timing more densely with two-checkpoint schedules and test whether PreviewDiff remains useful as the base model scale increases.

\section{Results}
We analyze PreviewDiff across all all image and video settings. Our multimodal critic models consist of Gemini-3.1-Flash-Lite and Qwen-3 when scoring denoised checkpoints. We also use Gemini-3.1-Flash-Lite and Qwen-3 as part of our evaluation, scoring the generated images/videos. We analyze these choices among others below. 

\begin{figure}
    \centering
    \includegraphics[width=\textwidth]{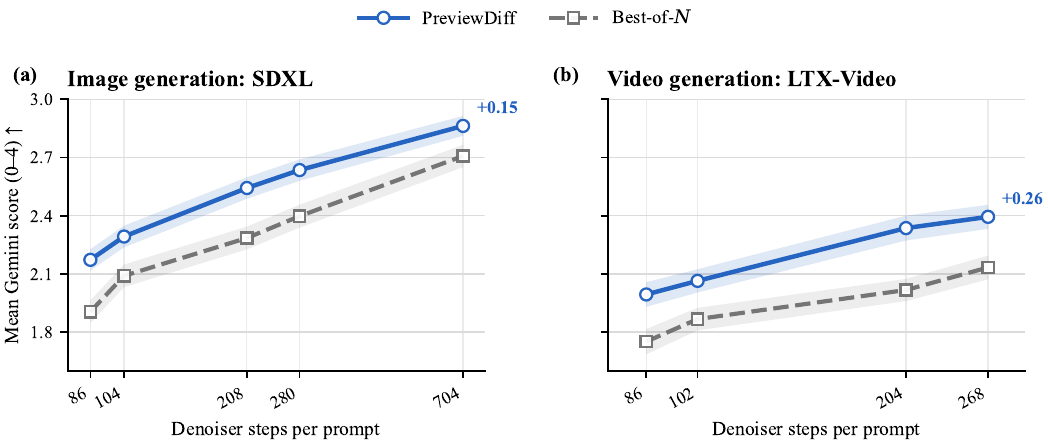}
    \caption{
    \textbf{PreviewDiff as a method for test-time scaling.}
    PreviewDiff improves with additional denoiser steps per prompt and outperforms budget-matched Best-of-$N$ selection for both image generation with SDXL and video generation with LTX-Video.}
    \label{fig:scaling}
\end{figure}

\begin{table}[t]
\centering
\caption{
Image generation results across image backbones and compositional benchmarks. 
}
\label{tab:image_results}
{\small
\setlength{\tabcolsep}{2.4pt}
\renewcommand{\arraystretch}{1.08}
\setlength{\sotatablewidth}{0.98\linewidth}
\begin{tabularx}{\sotatablewidth}{@{}L{1.75cm}L{2.45cm}YYYYYY@{}}
\toprule
\multirow{2}{*}{Model} 
& \multirow{2}{*}{Method}
& \multicolumn{3}{c}{GenEval2}
& \multicolumn{1}{c}{\shortstack{T2I-\\CoReBench}}
& \multicolumn{2}{c}{\shortstack{T2I-\\CompBench++}} \\
\cmidrule(lr){3-5}
\cmidrule(lr){6-6}
\cmidrule(l){7-8}
& 
& \shortstack{Gemini\\P@4}
& \shortstack{Qwen-3\\Val}
& \shortstack{Qwen-3\\P@4}
& \shortstack{Gemini\\P@4}
& \shortstack{Gemini\\P@4}
& \shortstack{BLIP\\VQA} \\
\midrule

\multirow{4}{*}{SDXL}
& PreviewDiff       & \bf 2.292 & \bf 1.9765 & \bf 1.975 & \bf 2.015 & \bf 3.474 & \bf 0.4844 \\
& Best-of-$N$       & 2.090 & 1.5523 & 1.735 & 1.713 & 2.963 & 0.3976 \\
& EvoSearch         & 2.103 & -- & 1.6196    & 1.7543    &   3.013  & 0.4521     \\
& Particle-Sample   & 1.902 & -- & 1.7544    & 1.9154    & 3.003    & 0.4255     \\

\midrule
\multirow{4}{*}{SD-3.5}
& PreviewDiff       & \bf 1.965 & \bf 1.588 & \bf 1.892    & \bf 1.765 & \bf 2.011 & \bf 0.2566 \\
& Best-of-$N$       & 1.754 & 1.3854 & 1.713  & 1.545 & 1.913 & 0.2011 \\
& EvoSearch         & 1.865 & -- & 1.812    & 1.712    &   1.975  & 0.1921     \\
& Particle-Sample   & 1.723 & -- & 1.809    & 1.701    & 1.865    & 0.2155     \\

\midrule
\multirow{4}{*}{\shortstack[l]{FLUX.2\\9B}}
& PreviewDiff           & \bf 3.112 & \bf 2.675 & \bf 1.983 & \bf 2.781 & \bf 3.758 & \bf 0.7851 \\
& Best-of-$N$       & 2.775 & 2.504 & 1.875 & 2.674 & 3.510 & 0.7256 \\
& EvoSearch         & 2.913 & -- &  1.923  &   2.752  &  3.601  & 0.7445     \\
& Particle-Sample   & 2.853 & -- & 1.920    & 2.730    & 3.456    & 0.7098     \\
\midrule
\multicolumn{8}{@{}l}{\footnotesize\itshape Published benchmark references using official benchmark metrics} \\[-1pt]
\sotarefrow
    {huang2025t2i}
    {T2I-CompBench++}
    {DaLLE-3}
    {BLIP-VQA}
    {0.6966}
\bottomrule
\end{tabularx}
}
\end{table}

\begin{table}[t]
\centering
\caption{
Video generation results across video backbones and text-to-video benchmarks.
Gray rows at the bottom show published benchmark reference results using each benchmark's official metric.
These rows are included for context and are not compute-matched to our methods. 
}
\label{tab:video_results}
{\small
\setlength{\tabcolsep}{2.4pt}
\renewcommand{\arraystretch}{1.08}
\setlength{\sotatablewidth}{0.98\linewidth}
\begin{tabularx}{\sotatablewidth}{@{}L{1.75cm}L{2.45cm}YYYYYY@{}}
\toprule
\multirow{2}{*}{Model} 
& \multirow{2}{*}{Method}
& \multicolumn{3}{c}{T2V-CompBench}
& \multicolumn{2}{c}{NarrLV}
& \multicolumn{1}{c}{V-Bench 2.0} \\
\cmidrule(lr){3-5}
\cmidrule(lr){6-7}
\cmidrule(l){8-8}
& 
& \shortstack{Gemini\\P@4}
& \shortstack{Qwen3-VL\\P@4}
& \shortstack{LLaVA\\Eval}
& \shortstack{Gemini\\P@4}
& \shortstack{Qwen-2.5\\Eval}
& \shortstack{Gemini\\P@4} \\
\midrule

\multirow{5}{*}{\shortstack[l]{LTX-Video\\9B}}
& MCTS-CGD          & {\bf 3.021} & {\bf 1.987} & {\bf 0.4523} & {\bf 2.167} & {\bf 62.17} & {\bf 3.162} \\
& Best-of-$N$       & 2.716 & 1.411 & 0.3922 & 1.544 & 55.92 & 3.123 \\
& EvoSearch         & 2.814 & 1.901 & 0.3713 & 2.012 & 61.14 & 3.071 \\
& Particle-Sample   & 2.755 & 1.926 & 0.3656 & 1.975 & 58.96 & 3.112 \\
& Video-T1          & 2.861 & 1.954 & 0.4329 & 2.115 & 61.87 & 3.124 \\

\midrule

\multirow{5}{*}{\shortstack[l]{Wan 2.2\\A14B}}
& MCTS-CGD          & {\bf 3.541} & {\bf 2.019} & {\bf 0.6311} & {\bf 3.411} & {\bf 71.22} & {\bf 3.554} \\
& Best-of-$N$       & 3.362 & 1.875 & 0.6011 & 2.874 & 68.16 & 3.292 \\
& EvoSearch         & 3.392 & 1.978 & 0.6126 & 3.012 & 68.75 & 3.399 \\
& Particle-Sample   & 3.386 & 1.944 & 0.5928 & 2.961 & 68.11 & 3.287 \\
& Video-T1          & 3.465 & 1.901 & 0.6266 & 3.229 & 70.02 & 3.509 \\

\midrule
\multicolumn{8}{@{}l}{\footnotesize\itshape Published benchmark references using official benchmark metrics} \\[-1pt]
\sotarefrow
    {hassan2025factorized}
    {T2V-CompBench}
    {Veo-3}
    {LLaVA-Eval}
    {0.6844}
\sotarefrow
    {hassan2025factorized}
    {T2V-CompBench}
    {Wan2.2}
    {LLaVA-Eval}
    {0.5742}
\sotarefrow
    {feng2025narrlv}
    {NarrLV}
    {Wan 2.2}
    {Qwen-2.5 Eval}
    {70.6}
\bottomrule
\end{tabularx}
}
\end{table}

\subsection{PreviewDiff improves test-time scaling}

Figure~\ref{fig:scaling} shows the main test-time scaling result. Across both image and video generation, PreviewDiff improves as the denoiser-step budget increases and remains above budget-matched Best-of-$N$ at every tested budget. On SDXL, PreviewDiff improves from 2.17 to 2.86 mean Gemini score as compute increases, while Best-of-$N$ rises from 1.91 to 2.71. On LTX-Video, PreviewDiff improves from 1.99 to 2.39, while Best-of-$N$ improves from 1.75 to 2.13. The gap persists at both small and large budgets, which suggests that intermediate feedback is not only a low-budget shortcut. It remains useful even when Best-of-$N$ is allowed to draw many complete samples.

This result supports the central premise of PreviewDiff. Final-output selection can only choose among completed samples. PreviewDiff can diagnose a partial generation, branch over targeted corrections, and allocate later computation to states that are already moving toward the prompt. The same verifier is therefore used more effectively when it guides the denoising process rather than only ranking the final outputs.

\subsection{Fixed-budget results across models and benchmarks}

Tables~\ref{tab:image_results} and~\ref{tab:video_results} evaluate PreviewDiff under fixed budgets across image and video backbones. We include evaluations with both open- and closed-source multimodal LLMs like Gemini-3.1-Flash-Lite and Qwen-3 as well as include basic evaluations from a particular benchmark like evaluating with BLIP-VQA for T2I-CompBench++. On image generation, PreviewDiff improves over Best-of-$N$ on most reported metrics across SDXL, SD-3.5, and FLUX.2. The gains are clearest on GenEval2, where the task stresses the same count, attribute, and relation errors that the multimodal critic is designed to identify.

PreviewDiff also improves auxiliary evaluator scores in the current table, including Qwen-3 (P@4) and BLIP-VQA columns, which suggests that the method is not merely overfitting to the Gemini critic. We also evaluate using Qwen-3 as a critic in PreviewDiff and Best-of-$N$ rather than as only an evaluator (Qwen-3 Val). We find significant improvement in these settings as well. This indicates that we can use Qwen-3 to provide consistent feedback. We note that EvoSearch and Particle Sampling do not use critic feedback from a multimodal judge so these are intentionally left blank.

On video generation, PreviewDiff improves over Best-of-$N$ on LTX-Video and Wan 2.2 across the main Gemini evaluations. The current LTX-Video results show especially large gains on T2V-CompBench and NarrLV, where temporal relations and prompt satisfaction are difficult to recover through final selection alone. Wan 2.2 shows the same pattern at a stronger base-model scale. We also find that using corresponding open-source models like Qwen-3, Qwen-2.5 or LLaVA as an evaluator achieves similar improvement, meaning we don't overfit to Gemini results. 

We find PreviewDiff outperforms other test-time scaling baselines like EvoSearch, Particle Sampling, and Video-T1. This indicates that using denoised trajectories as part of our evaluation is important to our improvements. Similarly, having an MLLM in the loop allows us to find better trajectories than what is proposed in EvoSearch. 
\subsection{Qualitative behavior}
\begin{figure}
    \centering
    \includegraphics[width=0.75\textwidth]{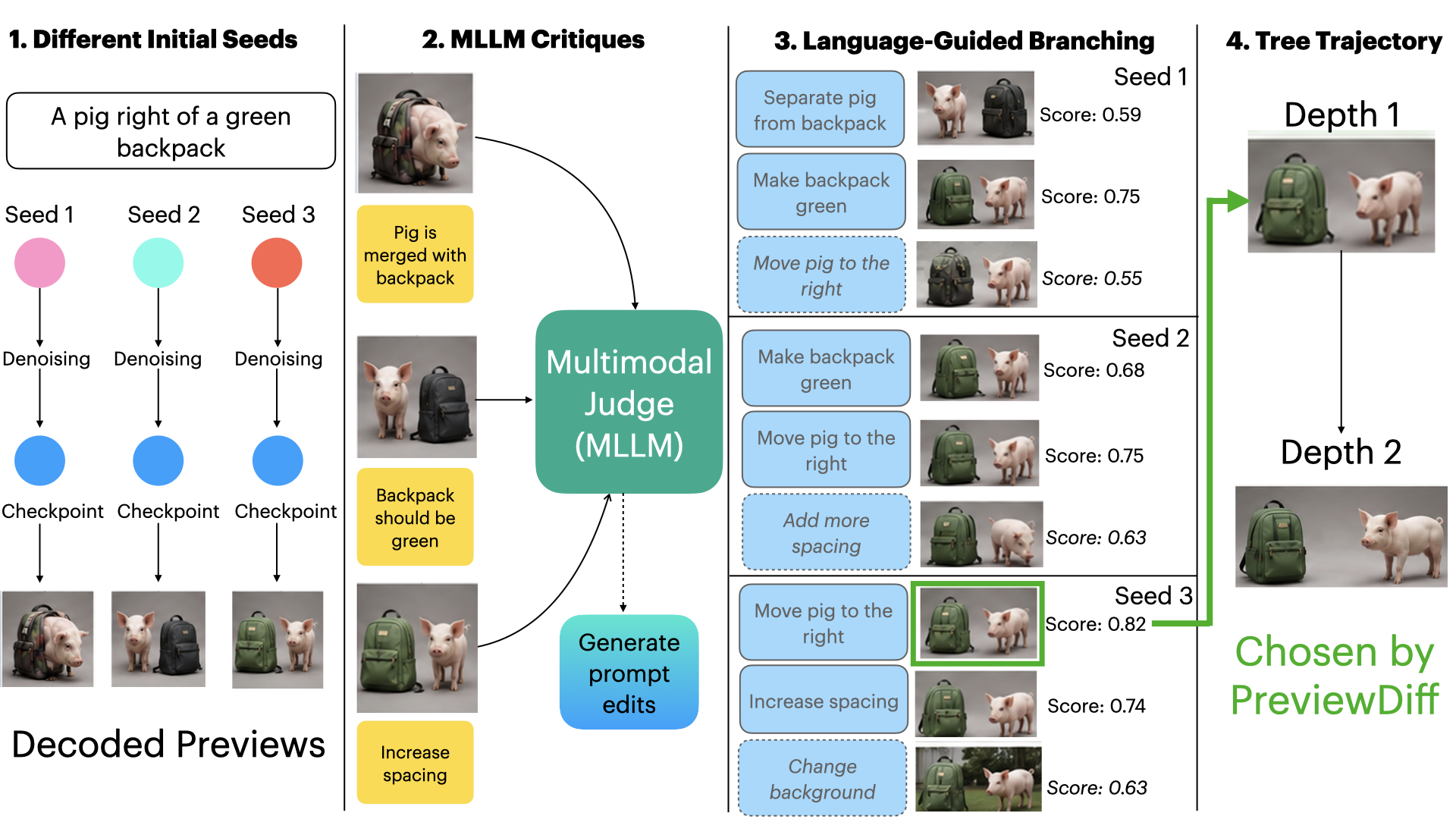}
    \caption{
    \textbf{PreviewDiff generation walkthrough.}
    Qualitative walkthrough of PreviewDiff from prompt to the chosen generation. 
    Our methods starts from a prompt and starts with several random seeds to get initial denoised images from a particular checkpoint. Due to the stochastic nature of diffusion models, this leads to different images. A multimodal judge analyzes the images and provides feedback (in the yellow boxes) and generates prompt edits. The summarized prompt edits for this example are shown in the third column with the new generated image and score from another multimodal judge. This score is between 0 and 1, only measuring prompt adherence. The final branch is chosen based on the highest score for each seed. We only show the tree trajectory for one image.} 
    \label{fig:image_qual}
\end{figure}

Figure~\ref{fig:image_qual} provides a qualitative analysis of the mechanisms used in PreviewDiff. For our generations, the most common successful corrections involve missing objects, wrong counts, wrong colors, and spatial relation errors. These examples show both the decoded preview and the branch selected by the critic, so we can see that PreviewDiff is not simply sampling more outputs. It is using intermediate visual evidence to steer the continuation.

We can observe that many corrections in PreviewDiff involve change object order, introducing spacing to aid visual style, or correcting physically implausible generations. From our analysis, language guidance is often correctly able to identify visual problems in the generated image as branching occurs. For example, an initial generation can be physically implausible e.g., the prompt requests a pig to the right of a green backpack and the resulting image merges and pig and backpack together. In the example, we find that MLLM, Gemini-3.1-Flash can identify these visual issues. Color issues are also common mistakes in accurately showing content. MLLM judges are also sensitive about spacing between objects, which is the most common stylistic complaint.

\subsection{Human Evaluation}
\begin{figure}
    \centering
    \includegraphics[width=0.75\textwidth]{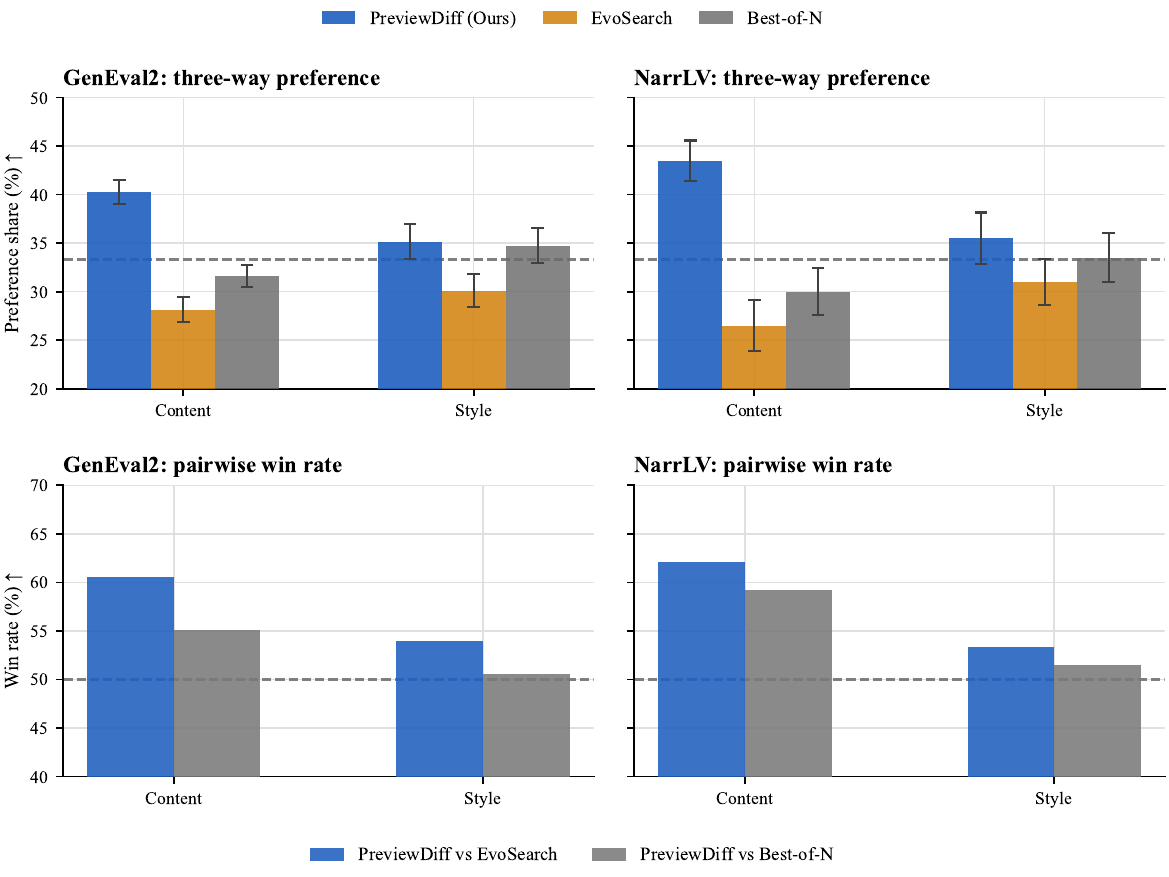}
    \caption{\textbf{Human evaluation of PreviewDiff}. We include a human evaluation of PreviewDiff, comparing with EvoSearch and Best-of-N. We demonstrate that PreviewDiff has strong improvement in accurately visualizing content in our generated images and videos, consistently beating other methods like Best-of-N and EvoSearch. While style gains are weaker, our strongest finding is that PreviewDiff significantly improves the content of the image or video.}
    \label{fig:human_eval}
\end{figure}

To further support our findings with MLLM evaluators, we include a human evaluation where we compare PreviewDiff against EvoSearch and Best-of-N, rating based on style and content. We show results in Figure~\ref{fig:human_eval}. We see that in all cases, PreviewDiff beats the other baselines including EvoSearch and Best-of-N. We find that PreviewDiff has the highest three way win rate compared to both methods, performing significantly above chance. Similarly, in head-to-head comparisons, PreviewDiff wins significantly more than EvoSearch and Best-of-N. 

When doing further analysis, we find that we improve significantly with content over style. The human results show larger gains for content than for style. This is expected because the PreviewDiff critic is prompted to focus on prompt adherence, including objects, attributes, counts, etc. The search therefore directly optimizes semantic correctness more than aesthetic quality. Style improvements are still possible when better semantic corrections also improve visual coherence, but they are not the primary target of the current critic. SDXL and LTX-Video have weaker performance style-wise. Improvements over style will likely happen with more robust underlying models. 

\subsection{Ablations on search axes}
\begin{figure}
    \centering
    \includegraphics[width=0.8\textwidth]{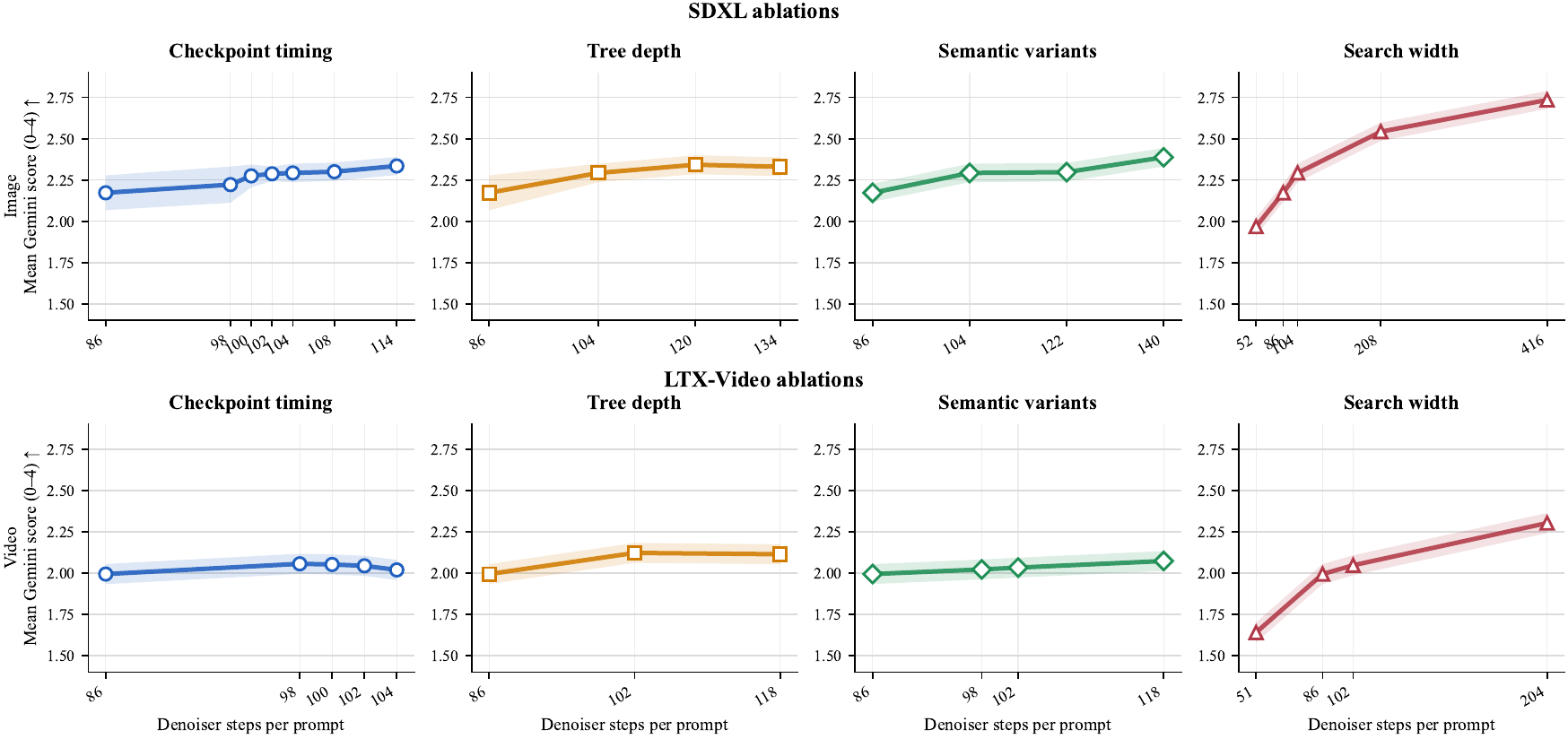}
    \caption{
    \textbf{Scaling across different search axes.}
    We ablate four axes of PreviewDiff for image generation with SDXL and video generation with LTX-Video, plotting mean Gemini score against total denoiser steps per prompt. 
    \emph{Checkpoint timing} varies the denoising step at which the first intermediate preview is decoded and judged. We vary our checkpoints from 2 checkpoints to 8 checkpoints. 
    \emph{Tree depth} varies the number of sequential judge-and-branch rounds. We vary from a depth of 1 to a depth of 3. 
    \emph{Semantic variants} varies the number of candidate prompt edits proposed by the multimodal judge at each expansion. We sweep from 1 variant to 4 variants. 
    \emph{Search width} varies the number of trajectories retained and explored. We sweep the width from 1 branch to 8 branches. 
    }
    \label{fig:ablations1}
\end{figure}

Figure~\ref{fig:ablations1} isolates the main axes of PreviewDiff. Search width produces the largest gains for both images and videos. On SDXL, increasing width raises the mean score from 1.97 to 2.74. On LTX-Video, width raises the score from 1.64 to 2.30. This is expected because width increases both the diversity of root trajectories and the number of partial states that survive pruning. It gives the critic more chances to find a promising continuation before the sample is finalized.

Depth and semantic variants provide smaller but useful gains. Increasing depth allows the method to correct a generation more than once, although returns saturate once the search reaches later checkpoints. Increasing semantic variants gives the critic more possible fixes to try at each node, but the benefit depends on whether the preview error is specific enough for the critic to propose distinct corrections. Checkpoint timing is also important. Earlier checkpoints leave more room for the corrected conditioning to affect the sample, while later checkpoints provide clearer previews but less opportunity to change the final output.

\subsection{Checkpoint timing and model scale}
\begin{figure}
    \centering
    \includegraphics[width=0.96\textwidth]{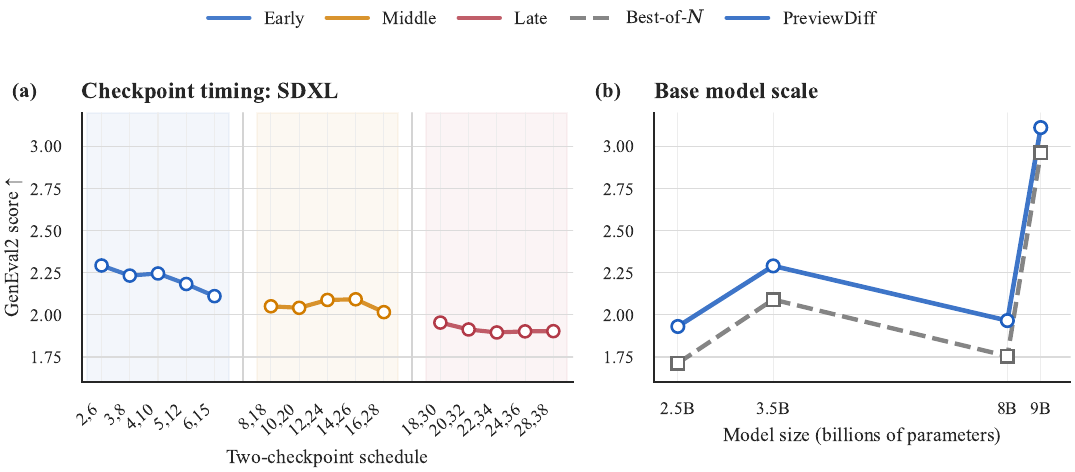}
    \caption{
    \textbf{Checkpoint timing and base-model scaling.}
    \textbf{(a)} Effect of checkpoint timing on PreviewDiff for SDXL on GenEval2.
    Each point corresponds to a two-checkpoint schedule $(c_1,c_2)$ indicating the denoising steps at which intermediate previews are decoded and judged.
    \textbf{(b)} Scaling with base model size, comparing budget-matched Best-of-$N$ selection against PreviewDiff at a fixed test-time budget.
    }
    \label{fig:checkpoint_scale}
\end{figure}

Figure~\ref{fig:checkpoint_scale} studies two complementary questions. First, it asks when the critic should intervene. Early two-checkpoint schedules consistently outperform middle and late schedules on SDXL. This indicates that intermediate previews do not need to be perfect to be useful. They only need to reveal enough semantic structure for the critic to identify a plausible correction. Intervening too late gives the judge a clearer preview, but leaves too little denoising time for the correction to reshape the sample.

Second, Figure~\ref{fig:checkpoint_scale} asks whether PreviewDiff is only useful for weaker base models. Across the tested model scales, PreviewDiff improves over fixed-budget Best-of-$N$. The gains remain present at the largest model size, where the base model already has stronger prompt-following ability. This suggests that critic-guided search is complementary to model scaling. Larger models produce better trajectories, but intermediate feedback still helps allocate test-time compute to the most promising continuations.

\subsection{Additional ablations}

\begin{table}[t]
\centering
\caption{
Component ablations for PreviewDiff.
Image results use SDXL on GenEval2, and video results use LTX-Video on NarrLV.
Initial-root ablations change the default two-root setting.
}
\label{tab:component_ablations}
{\small
\setlength{\tabcolsep}{3.2pt}
\renewcommand{\arraystretch}{1.08}
\begin{tabularx}{0.82\linewidth}{@{}L{4.6cm}YY@{}}
\toprule
\multirow{2}{*}{Experiment}
& \multicolumn{1}{c}{Image}
& \multicolumn{1}{c}{Video} \\
\cmidrule(lr){2-2}
\cmidrule(l){3-3}
&
\shortstack{SDXL on GenEval2\\Gemini P@4}
&
\shortstack{LTX-Video on NarrLV\\Gemini P@4} \\
\midrule
PreviewDiff
& {\bf 2.292}
& {\bf 2.167} \\
\hspace{1.1em}w/o language feedback
& 2.063
& 2.014 \\
\hspace{1.1em}w/o prompt editing
& 2.285
& 2.125 \\
\hspace{1.1em}initial roots $2 \rightarrow 1$
& 2.290
& 2.156 \\
\hspace{1.1em}initial roots $2 \rightarrow 3$
& 2.310
& 2.178 \\
\bottomrule
\end{tabularx}
}
\end{table}

In Table~\ref{tab:component_ablations}, we ablate specific components of PreviewDiff. PreviewDiff can be modified to stop using language feedback as part of the generation process, so the multimodal judge only provides a numerical score. Language feedback provides useful guidance for ensuring generation quality across the tree search using intermediate denoised checkpoints. Similarly, we test whether prompt editing via the feedback from the multimodal judge improves results. Similar to findings in Figure~\ref{fig:checkpoint_scale}, prompt editing to add or change the prompt doesn't have very strong effects on the final generation quality but does cause an improvement. Finally, the number of initial sampled roots can affect the final generation. We could start with multiple noised latents as the roots of our tree in MCTS. We find that this has little effect on our final result. Using 1, 2, or 3 nodes doesn't have a huge effect on our final result in comparison to branching. 

\section{Conclusion}
We introduced PreviewDiff, a training-free test-time search method that uses multimodal feedback inside the denoising process rather than only after generation is complete. By decoding intermediate latent previews, asking a multimodal judge to critique them, and branching over semantic prompt edits with local latent continuations, PreviewDiff turns visual generation into an adaptive closed-loop search. Across image and video benchmarks, this strategy improves over final-sample selection and scalar-search baselines under matched compute budgets. The results suggest that multimodal judges are most useful when they act not only as verifiers, but as semantic controllers that help decide where generation should go next.

\bibliography{main}
\newpage
\appendix
\section{PreviewDiff Prompting}
\label{ap:prompts}
\begin{figure}
    \centering
    \includegraphics[width=0.6\textwidth]{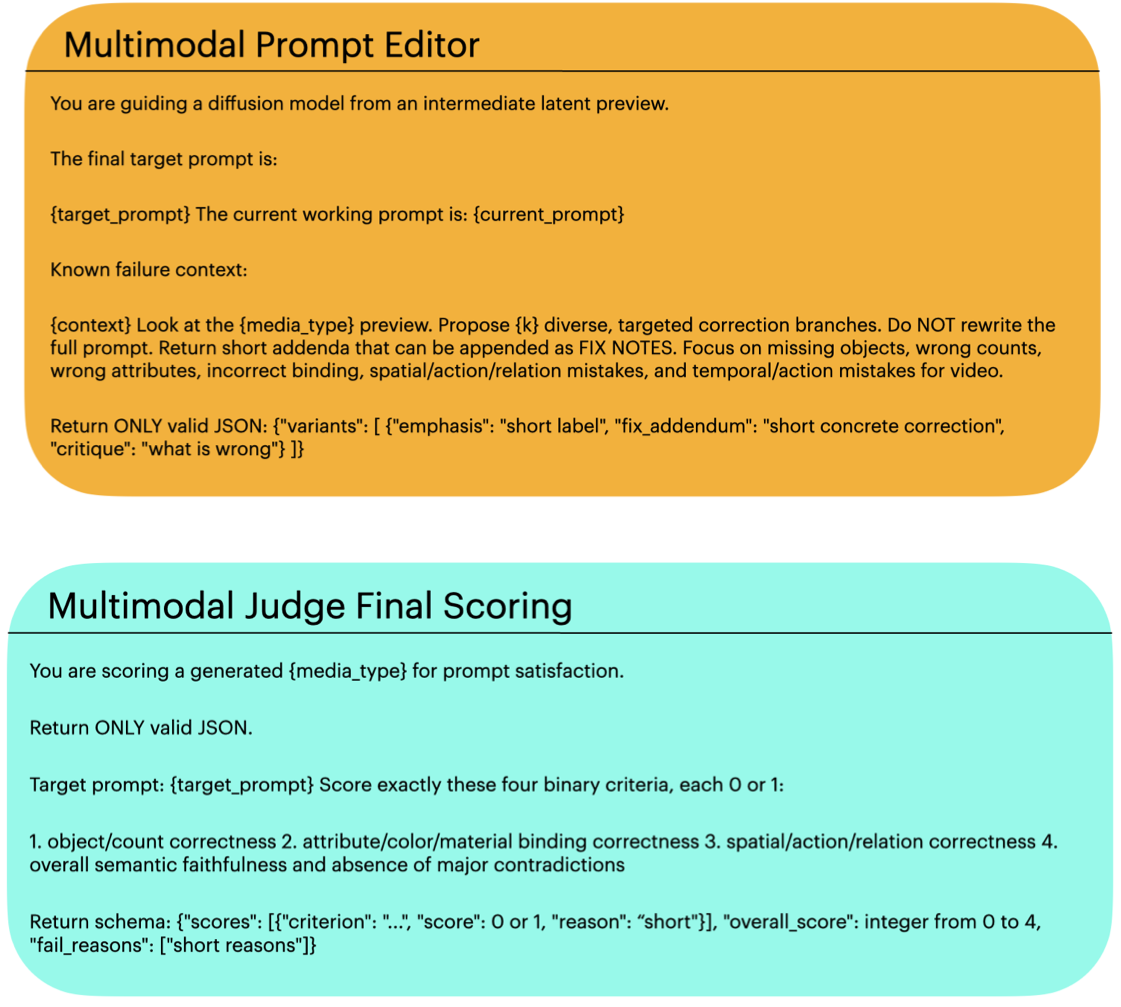}
    \caption{\textbf{Prompts used in PreviewDiff}. We include prompts used for our multimodal judge in PreviewDiff. The top prompt in orange shows the prompt edits applied by a multimodal judge over the tree search. Similarly, final scoring happens using a 0-4 scale, shown in the bottom in blue.}
    \label{fig:prompts}
\end{figure}

We describe prompts below. Specific prompts for our multimodal judge are included in Figure~\ref{fig:prompts}.

\paragraph{Diffusion Model Prompt}
The prompt fed to the image or video diffusion model is consist across different base generators. Given the original prompt \(p\) and a list of recent critic fix notes \(\Delta c_1,\ldots,\Delta c_m\), the positive conditioning prompt is
\[
p^+ = p \oplus \texttt{``FIX NOTES: ''} \oplus \Delta c_{m-1} \oplus \Delta c_m .
\]
When negative prompting is available, we append recent critique fragments to the negative prompt using the prefix \texttt{``Avoid:''}.

\paragraph{Final evaluator prompt.}
PreviewDiff uses the same multimodal judge for final image and videos. The judge receives the target prompt and the final output. It returns a JSON object with four binary criteria: object/count correctness, attribute/color/material binding, spatial/action/relation correctness, and overall semantic faithfulness. The sum gives a score from 0 to 4.

\paragraph{Search-time prompt.}
PreviewDiff uses the a multimodal judge for previewed images and videos. The judge receives the target prompt and the final output. It returns score between 0 and 1, scoring the image on faithfulness to the prompt. This is done so we don't overfit to the final evaluator prompt. The prompt follows the final evaluator prompt but does not use a JSON with the four requested evaluation axes. 

\paragraph{Language-feedback prompt.}
For branch expansion, the judge is also asked to propose short correction addenda. These addenda are not full prompt rewrites. They are appended to the original user prompt under a \texttt{FIX NOTES} field, while critique fragments are optionally appended to the negative prompt using an \texttt{Avoid:} prefix. This keeps the original prompt fixed while allowing the next denoising continuation to focus on the specific error observed in the intermediate preview.

\section{Human Evaluation}
\label{ap:human_eval}
\begin{figure}
    \centering
    \includegraphics[width=\textwidth]{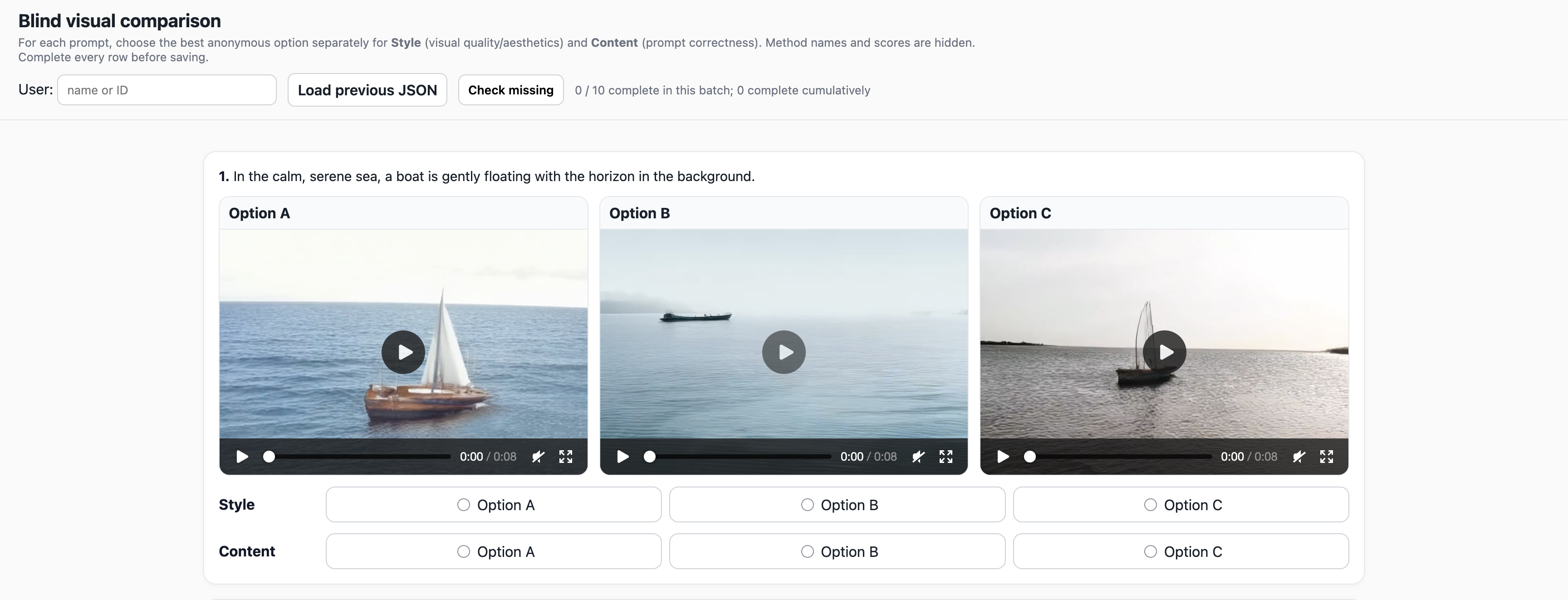}
    \caption{\textbf{Human Evaluation Page}. We provide an example of our human evaluation webpage, shown to participants for rating output images and videos. Our evaluation consists of two axes, style and content. Raters are asked to choose between the three videos based on both axes.}
    \label{fig:human_eval_page}
\end{figure}

To fully verify the improvement of PreviewDiff, we include a full human evaluation using nine human raters. Each human rater was asked to evaluate the image/video outputs of three methods for a single prompt. There were 360 image prompts, taken from GenEval2, and 200 video prompts, taken from NarrLV. 

To recruit human raters, we chose adults over the age of 18. All raters were unfamiliar with the nature of the work nor were the goals of the evaluation made clear to the human raters. Raters were requested to evaluate the generations for each prompt on style and content. The instruction stated: \texttt{For each prompt, choose the best anonymous option separately for Style (visual quality/aesthetics) and Content (prompt correctness, object correctness over background correctness, consistency, etc.).}

A screen capture of our human evaluation UI is included in Figure~\ref{fig:human_eval_page}. The UI demonstrates the labels are hidden from raters. Furthermore, raters were asked to choose for both style and content.

For the purpose of this evaluation, we did not include ties. We did so to get accurate human ratings given that style comparisons often very similar. PreviewDiff wasn't prompted to improve style ratings so many times, raters would choose ties. This complicated the ratings. 

\section{Visual Examples}
\label{ap:visual_examples}
\begin{figure}
    \centering
    \includegraphics[width=\textwidth]{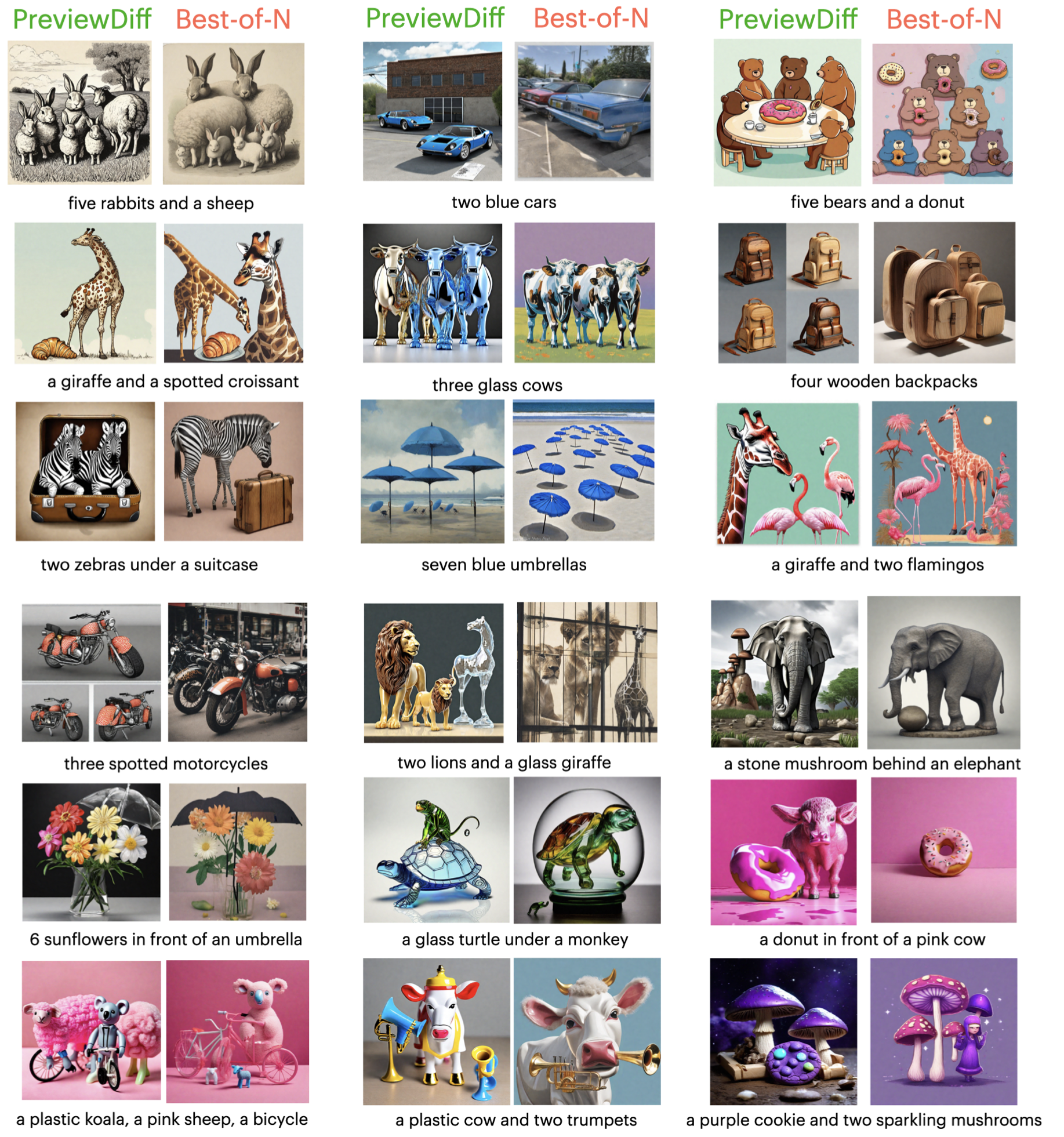}
    \caption{\textbf{Image Examples from PreviewDiff vs Best-of-N.} We include 18 text-to-image prompts comparing PreviewDiff vs Best-of-$N$. The base model is SDXL and the prompts are sampled from GenEval2 and T2I-CoReBench. Across all examples, using PreviewDiff leads to improved visual components with respect to style and content. For example, we correct numerical counts (five rabbits, two blue cars) and include corrections with texture (three glass cows). While numerical mistakes still occur in PreviewDiff, the errors are less severe than what is found in Best-of-$N$ (eight umbrellas vs >ten).}
    \label{fig:visual}
\end{figure}

\begin{figure}
    \centering
    \includegraphics[width=\textwidth]{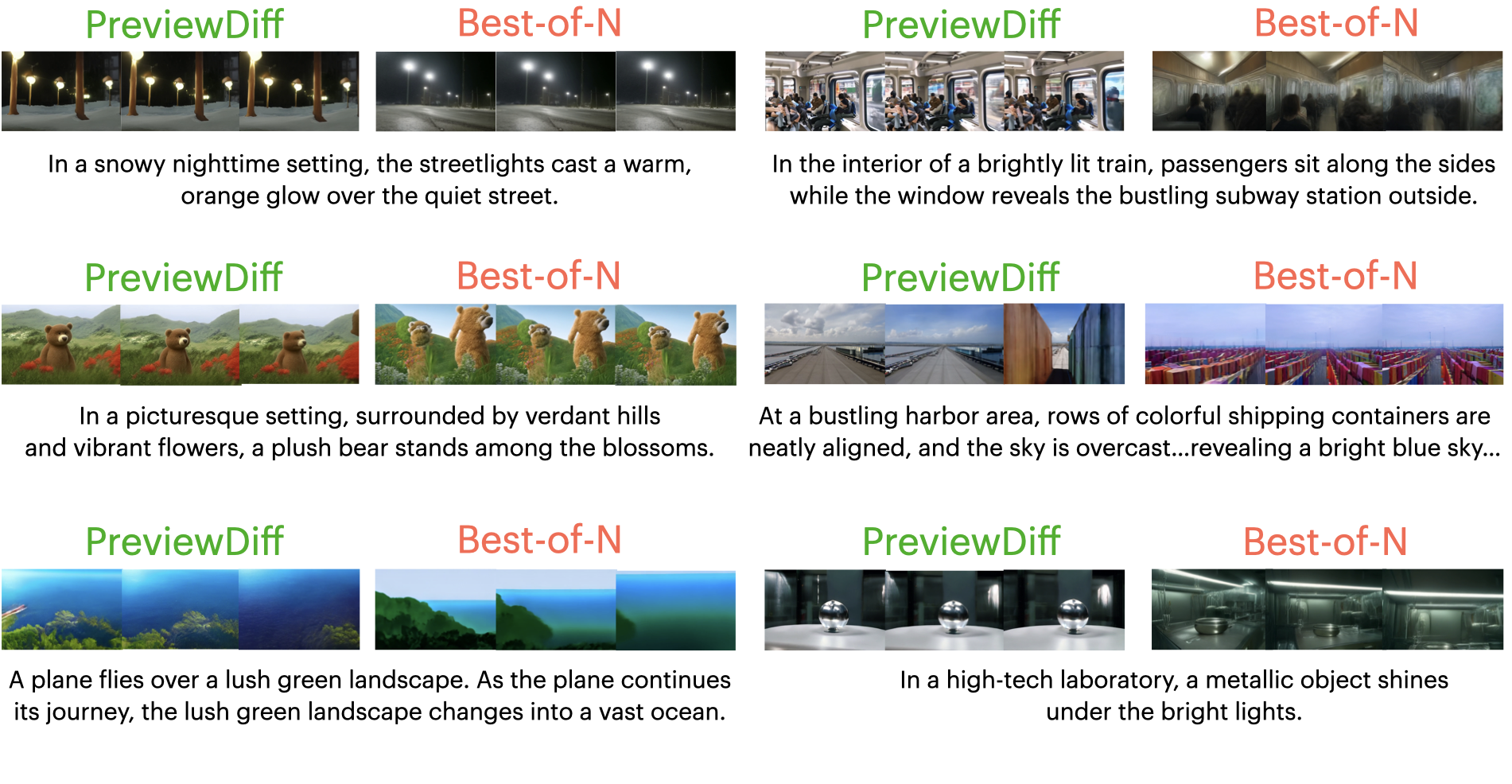}
    \caption{\textbf{Video Examples from PreviewDiff vs Best-of-$N$.} We include 6 text-to-video prompts comparing PreviewDiff vs Best-of-$N$. The base model is LTX-Video and the prompts are sampled from NarrLV. Across all examples, using PreviewDiff leads to improved visual components with respect to style and content. For example, we correct colorization (orange glow, brightly lit) and include corrections with lighting (shines under the bright lights). While mistakes still occur in PreviewDiff, the errors are less severe than what is found in Best-of-$N$ (e.g. harbor area is not bustling).}
    \label{fig:video}
\end{figure}

We show additional image examples in Figure~\ref{fig:visual} and video examples in Figure~\ref{fig:video}. These qualitative examples highlight a consistent pattern: Best-of-$N$ often selects a visually plausible sample, but the selected output may still miss a key prompt constraint. PreviewDiff instead uses the intermediate preview to identify the specific failure and steer the continuation toward a more faithful sample. For images, the improvements are most visible on compositional errors. PreviewDiff more often recovers the correct number of objects, fixes spatial relations such as objects being under, in front of, or to the left of another object, and improves attribute binding for colors and materials such as glass, stone, or wood. 
This is especially important on GenEval2, where many prompts are simple in language but difficult for open-loop sampling because the model must satisfy several object-level constraints at once.

For videos, the same mechanism improves prompt adherence over time.  Best-of-$N$ can produce videos with appealing motion or appearance, but it frequently leaves failures in scene color, lighting, object identity, or the intended action relation. PreviewDiff often corrects these issues by using the critic's language feedback to preserve promising structure while revising the parts of the trajectory that conflict with the prompt. The video examples show improvements in lighting and colorization, clearer depiction of the requested scene, and better alignment between the main subject and the described action or environment. Together, Figures~\ref{fig:visual} and~\ref{fig:video} illustrate that PreviewDiff is not simply selecting more attractive samples. It is using intermediate visual evidence to make targeted corrections, which leads to outputs that are both visually coherent and more faithful to the prompt.

\section{Failures Cases}
\begin{figure}
    \centering
    \includegraphics[width=\textwidth]{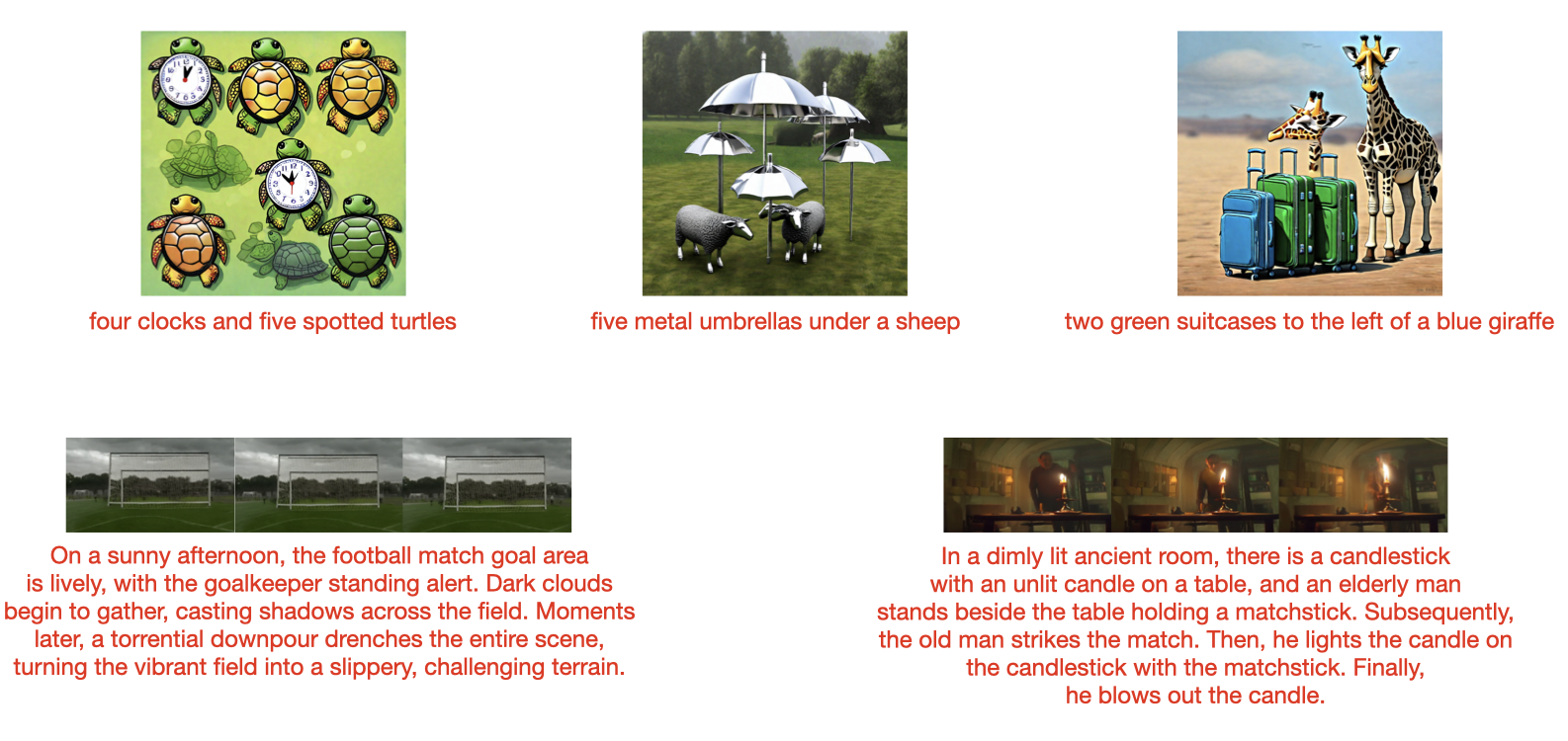}
    \caption{\textbf{PreviewDiff weaknesses and failures}. We show examples where PreviewDiff does not generate the correct result particularly with cases of numeracy, spatial relations, and attribute binding. }
    \label{fig:failures_cases}
\end{figure}

Although PreviewDiff improves prompt following across image and video generation, it does not eliminate all compositional failures. Figure~\ref{fig:failures_cases} shows representative failure modes. For images, the remaining errors often involve constraints that are discrete, relational, or highly specific. PreviewDiff may recover the right object categories while still missing exact numeracy, as in the prompt requiring four clocks and five spotted turtles. It can also invert spatial relations, such as depicting sheep under umbrellas rather than umbrellas under a sheep. Attribute binding remains imperfect as well: in the blue-giraffe example, the method preserves the requested layout but fails to bind the blue color to the giraffe.

For videos, failures are most common when the prompt requires a precise temporal sequence rather than a static scene. PreviewDiff can generate a plausible goal area but still omit the goalkeeper and the downpour event. It can also skip intermediate actions, such as showing a candle already lit rather than depicting the full sequence of striking a match, lighting the candle, and blowing it out. Finally, temporal color ordering remains difficult. In the crystal example, the object changes appearance, but not in the requested transparent-to-blue-to-green-to-yellow-to-orange-to-purple sequence. These cases suggest that PreviewDiff is strongest when semantic errors can be corrected by local prompt-guided continuation, but weaker when success requires exact counting, relation grounding, or long-horizon temporal state tracking.
\end{document}